%% file: root.tex
\documentclass[letterpaper, 10 pt, conference]{ieeeconf}

\IEEEoverridecommandlockouts

\usepackage{graphics}
\usepackage{epsfig}
\usepackage{mathptmx}
\usepackage{times}
\usepackage[fleqn]{amsmath}
\usepackage{amssymb}
\usepackage{listings}
\usepackage{algorithm}
\usepackage[noend]{algpseudocode}
\usepackage{xcolor}
\usepackage{subcaption}
\usepackage{siunitx}
\usepackage{import}
\usepackage{pgf}
\usepackage[utf8]{inputenc}
\usepackage{orcidlink}
\usepackage{balance}
\DeclareUnicodeCharacter{2212}{-}
\hypersetup{colorlinks=false,hidelinks=true}

\title{\LARGE \bf
Iterative Surface Mapping Using Local Geometry Approximation with Sparse
Measurements During Robotic Tooling Tasks
}
\author{
  Manuel Amersdorfer\orcidlink{0000-0002-6416-3453},
  \IEEEmembership{Student Member, IEEE}
  and
  Thomas Meurer\orcidlink{0000-0002-9175-2157},
  \IEEEmembership{Senior Member, IEEE}
  \thanks{The authors are with the Chair of Automatic Control, Faculty of
  Engineering, Kiel University, Kaiserstraße 2, 24143 Kiel, Germany,
  {\tt\small \{maam,tm\}@tf.uni-kiel.de}}
}

\input{abbreviations}

\usepackage[absolute,showboxes]{textpos}
\TPMargin{5pt}

\newcommand{\copyrightstatement}{
    \begin{textblock}{0.87}(0.06,0.93)    
        \noindent
        \textcopyright 2021 IEEE.  Personal use of this material is permitted.  Permission from IEEE must be obtained for all other uses, in any current or future media, including reprinting/republishing this material for advertising or promotional purposes, creating new collective works, for resale or redistribution to servers or lists, or reuse of any copyrighted component of this work in other works.
    \end{textblock}
}

\begin{document}

\maketitle
\copyrightstatement
\thispagestyle{empty}
\pagestyle{empty}

\input{sections/0_abstract}
\input{sections/1_introduction}
\input{sections/2_surface_approximation}
\input{sections/3_surface_mapping}
\input{sections/4_experimental_setup}
\input{sections/5_results_and_discussion}
\input{sections/6_conclusion}

\balance

\section*{ACKNOWLEDGMENT}
The first author (Manuel Amersdorfer) is supported by the InProReg project
(project no. DD01-004). InProReg is financed by Interreg Deutschland-Danmark
with means from the European Regional Development Fund.

\bibliographystyle{IEEEtran}
\bibliography{literature}

\end{document}

%% file: abbreviations.tex
\renewcommand{\vec}[1]{{\pmb{#1}}}

\newcommand{\transp}{^\textrm{\scriptsize{T}}}

\newcommand{\dir}{\vec{e}}

\newcommand{\point}{\vec{p}}

\newcommand{\zEst}{\hat{z}}
\newcommand{\zReal}{\bar{z}}
\newcommand{\zError}{\tilde{z}}
\newcommand{\zMeasurement}{z}

\newcommand{\covApprox}{R}
\newcommand{\covProcess}{Q}
\newcommand{\covState}{\hat{P}}
\newcommand{\kalmanGain}{K}
\newcommand{\innovationCov}{S}
\newcommand{\mask}{M}

\newcommand{\radius}{r}

%% file: sections/0_abstract.tex
\begin{abstract}
We present a cost-efficient and versatile method to map an unknown 3D freeform
surface using only sparse measurements while the end-effector of a robotic
manipulator moves along the surface.
The geometry is locally approximated by a plane, which is defined by measured
points on the surface.
The method relies on linear Kalman filters, estimating the height of each
point on a 2D grid.
Therefore, the approximation covariance for each grid point is determined using
a radial basis function to consider the measured point positions.
We propose different update strategies for the grid points exploiting the
locality of the planar approximation in combination with a projection method.
The approach is experimentally validated by tracking the surface with a robotic
manipulator.
Three laser distance sensors mounted on the end-effector continuously measure points on the surface during the motion to determine the approximation plane.
It is shown that the surface geometry can be mapped reasonably accurate with a
mean absolute error below 1~mm.
The mapping error mainly depends on the size of the approximation area
and the curvature of the surface.
\end{abstract}

%% file: sections/1_introduction.tex
\section{INTRODUCTION}
\label{sec:introduction}
The knowledge of the exact geometry of a part is crucial for most automated
manufacturing techniques like milling, polishing or grinding.
In usual production scenarios, the geometry is determined by CAD drawings.
But in cases where this data is not available, it has to be reconstructed by
using highly accurate coordinate measuring machines (CMM) \cite{ren_curve_2017}
or vision-based techniques such as structured light sensing
\cite{hansen_structured_2014,hennad_characterization_2019,
song_distortion-free_2019}.
However, these techniques are expensive, require exhaustive calibration and can
only map a limited volume.
For the machining of large parts or mobile manufacturing scenarios, this may be
inappropriate.
Especially in small and medium-sized enterprises with high-mix low-volume
production, the required large investments for measurement equipment might
become an obstacle for the automation of manufacturing processes.
Therefore, it is our aim in this contribution to develop a versatile and
cost-efficient method to map an unknown geometry focusing on robotic tooling
tasks.
The resulting 3D model provides the basis to improve the path and motion
planning of the process.

The work presented in \cite{chu_visible_1988} uses local parameterized patches
from sparse depth measurements to reconstruct a surface on a dense 2D grid.
Therefore, polynomial surface patches are determined such that the maximum
approximation error is minimized.
In \cite{zhao_development_2008}, a 6-DOF robotic manipulator moves a part in
front of a stationary 2D laser scanner.
The scanned points are combined with the robot's forward kinematics to
reconstruct the geometry of the part.
This limits the approach to applications, where the part is small enough to be
moved by the robot. %
The geometry of the surface can also be determined during force-controlled
interaction between tool and surface, assuming a known contact point, as
demonstrated in \cite{ganesh_versatile_2012}.
However, this gives only the contour along the performed path but not in its
vicinity.
A similar approach is used in \cite{song_artistic_2018} with an impedance
controller to iteratively update an estimate of the surface geometry.
This requires an initial coarse sampling of the surface, which is represented as
bi-linear interpolation of the sampling points.
Other tactile approaches, such as those proposed in
\cite{mazzini_tactile_2009,mazzini_tactile_2011}, require geometric primitives
to identify the shape parameters from sparse measurement points.
In \cite{ozog_real-time_2013}, planar patches, obtained from sparse point
clouds, are used to reconstruct a large-scale surface.
These planar patches serve as features in a real-time SLAM implementation.
A similar approach is used in \cite{hong_three-dimensional_2019} for visual
mapping.
Kalman filter-based methods are widely used for SLAM but can also be exploited
for map data fusion \cite{slatton_fusing_2001}.
A probabilistic representation of an object model is proposed in
\cite{faria_probabilistic_2010}, where better explored regions have a higher
probability.

This paper presents a novel method to map the surface geometry using sparse
sensor measurements to approximate the surface locally.
The sensors are mounted on a robotic manipulator that moves along the surface.
During this motion, the mapping is continuously updated by these approximations.
In this way, a large area can be covered.
Another advantage of the robotic setup is that hidden parts can be mapped, which
would not be visible to stationary 3D scanners and the part is scanned from
different directions without requiring a rotary table.
By weighting the updates based on a radial basis function (RBF) as a heuristical
estimate of the approximation covariance, the mapping is iteratively improved.
Different region-based update schemes, implemented by binary masks,
allow us to update the mapping only in regions where the approximation is
valid. The approach is evaluated on an experimental setup.
The measurement works with laser distance sensors but can also be used with
other distance sensor types, e.g., tactile, inductive or capacitive, to measure
even materials which might be hard to detect with optical sensors, such as
transparent, reflective or dark surfaces.
Therefore, the approach provides an alternative solution to commercial 3D
scanning methods for applications where not their full $\si{\micro\m}$-accuracy
is required.

The paper is structured as follows:
First, the representation and the local approximation of the surface are
described in Section \ref{sec:surface_approximation}.
The mapping method is then introduced in Section \ref{sec:surface_mapping} by
defining the Kalman filters for height estimation, the measurement covariance
and some mask types for the local update.
Section \ref{sec:experimental_setup} describes the experimental setup, which is
used for the mapping of the example surface.
The results are presented and discussed in Section
\ref{sec:results_and_discussion}.
Section \ref{sec:conclusion} concludes the paper and gives some ideas for
future work.

%% file: sections/2_surface_approximation.tex
\section{SURFACE APPROXIMATION}
\label{sec:surface_approximation}
\begin{figure}
  \vspace{1.5mm}
  \centering
  \def\svgwidth{75mm}
  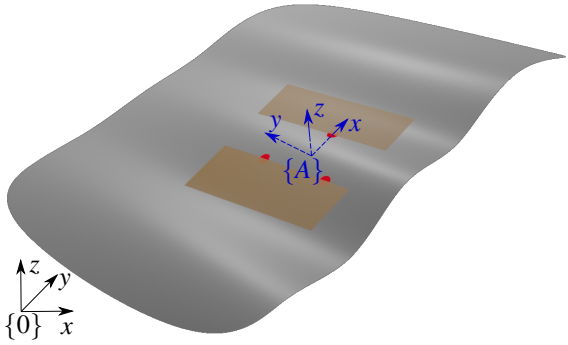
  \caption{Local approximation of the freeform surface (grey) as a plane
  (orange) determined from three measured points (red) and coordinate frames
  $\{ 0 \}$ (black) and $\{ A \}$ (blue).
  }
  \label{fig:curvecd_surface_approx}
\end{figure}
The proposed approach relies on the fact that the surface can be approximated
locally by a geometric primitive such as a plane.
In this section, we first define a mathematical representation of the surface,
which is used for the later mapping.
Afterwards, the local approximation of the geometry is described using sparse
measurement points on the surface.

\subsection{Surface Representation}
\label{subsec:surface_representation}
To represent the surface geometry, we define the surface as a height
map on a 2D grid.
Without loss of generality, the grid is given inside a rectangular area in the
$xy$-plane of the inertial coordinate system $\{ 0 \}$.
Therefore, the $x$- and $y$-coordinates are discretized in the intervals $x \in [x_\text{min}, \, x_\text{max}]$ and $y \in [y_\text{min}, \, y_\text{max}]$.
The geometry of the surface is determined by its height
\begin{subequations} \label{eq:surface_representation}
\begin{align}
  \zReal_{ij} &= f(x_i, \, y_j) \label{eq:surface_height}
  \intertext{on each grid point}
  x_i &= h_x i + x_\text{min}
  \quad i = 0, \, \ldots, \, N - 1 \label{eq:grid_x}\\
  y_j &= h_y j + y_\text{min}
  \quad j = 0, \, \ldots, \, M - 1, \label{eq:grid_y}
\end{align}
\end{subequations}
where the values $h_x = \frac{x_\text{max} - x_\text{min}}{N}$ and $h_y =
\frac{y_\text{max} - y_\text{min}}{M}$ denote the step size of the grid in $x$-
and $y$-direction.
The mapping $f: \mathbb{R}^2 \rightarrow \mathbb{R}$ is assumed to be unique
over the grid.
Using an affine transformation $\point' = (R_0^\text{A} )\transp (\point -
\point_0^\text{A})$ allows us to transform the desired part of the surface such
that it can be parameterized by the $xy$-plane of the local coordinate system
$\{ \text{A} \}$, where $R_0^\text{A} \in \mathrm{SO}(3)$ denotes the rotation
matrix and $\point_0^\text{A} \in \mathbb{R}^3$ the origin of the local
coordinate system with respect to the inertial coordinate system.

\subsection{Local Geometry Approximation}
\label{subsec:local_geometry_approximation}
As described in \cite{amersdorfer_real-time_2020}, the surface geometry can be
approximated using sparse measurements of points on the surface.
In Fig. \ref{fig:curvecd_surface_approx}, the local approximation of the
surface is shown.
Three measurement points are used to define a plane, which approximates
the surface geometry locally.
This provides a first-order approximation of the local geometry, which can be
expressed by the surface normal vector $\vec{n} = [n_x \: n_y \: n_z]\transp$
and the distance $p$ of the plane from the origin.
For $L \geq 3$ distinct measurement points $\point_1, \, \ldots, \, \point_L \in
\mathbb{R}^3$, the surface normal vector $\vec{n}$ can be determined as the
eigenvector corresponding to the smallest eigenvalue of the covariance matrix
\begin{equation}
  C =  \frac{1}{L} \sum_{l = 1}^L (\point_l - \point_c )
  (\point_l - \point_c)\transp
  \; \; \text{with} \;
  \point_c = \frac{1}{L} \sum_{l = 1}^L \point_l,
  \label{eq:measurement_points_covariance}
\end{equation}
where $\point_c$ is the centroid of all points.
This method is also known as principal component analysis (PCA) for point
clouds \cite{mitra_estimating_2003,fransens_hierarchical_2006}.
The distance $p$ of the plane from the origin then follows as
$p = \vec{n} \cdot \point_c$,
assuming that the plane goes through the center point $\point_c$.
For $L = 3$, all three measurement points and the center point lie on the plane.
The length of the normal vector equals to $\| \vec{n} \| = 1$ per definition.
The $z$-coordinate of the points on the approximation plane
$\point_0^S(x_i, \, y_j) = [x_i \: y_j \: \zMeasurement_{ij}]\transp$
is obtained as
\begin{equation}
  \zMeasurement_{ij}
  = \frac{1}{n_z} (p - n_x x_i - n_y y_j )
  \label{eq:z_measurement}
\end{equation}
for a given $x_i$ and $y_j$ value, assuming $n_z \neq 0$.

Notice that the approximation plane may intersect the surface depending on the
alignment of the measurement points.
According to the mean value theorem, this plane also approximates the tangential
plane of the surface.
In a sufficiently small area, the error between the surface geometry and the
approximation function is bounded.
This only holds as long as the curvature in this area is sufficiently small,
which is valid when the curvature radius is much larger than the distance
between the measurement points.
In this paper, it is assumed that the surface is approximated on a subset of the
grid points inside the approximation area $\mathcal{A} \subset \mathbb{R}^2$.
Let $\mathcal{I} := \lbrace(i, \, j) \vert (x_i, \, y_j) \in \mathcal{A},
\, \forall \, (i, \, j) \in \{1, \, \ldots, \, M\}
\times \{1, \, \ldots, \, N\} \rbrace$
be the set of indices for all grid points inside this area.
The definition of such a local approximation area is discussed in
Section \ref{subsec:masked_update}.

To project coordinates from the inertial coordinate system $\{ 0 \}$ onto the
plane, we define a local coordinate frame $\{ \text{A} \}$ on the plane.
The approximation center point is chosen as the origin of the frame $\point_0^A
= \point_c$.
The rotation matrix is defined by using the surface normal direction $\dir_z = \vec{n}$ as $z$-direction and a vector on the surface
\begin{subequations}
  \label{eq:rot_mat_local_frame}
  \begin{align}
    \dir_x &= \frac{\point_1 - \point_c - (\point_1 - \point_c)\cdot \vec{n} \, \vec{n}}{\| \point_1 - \point_c - (\point_1 - \point_c)\cdot \vec{n} \, \vec{n} \|} \\
    R_0^A &= \begin{bmatrix} \dir_x & \vec{n} \times \dir_x & \vec{n}  \end{bmatrix},
  \end{align}
\end{subequations}
where $\dir_x$ is the direction of the distance vector $\point_1 - \point_c$
projected onto the approximation plane, satisfying $\dir_x \perp \vec{n}$.

%% file: figures/png/curvecd_surface_approx.pdf_tex
\begingroup%
  \makeatletter%
  \providecommand\color[2][]{%
    \errmessage{(Inkscape) Color is used for the text in Inkscape, but the package 'color.sty' is not loaded}%
    \renewcommand\color[2][]{}%
  }%
  \providecommand\transparent[1]{%
    \errmessage{(Inkscape) Transparency is used (non-zero) for the text in Inkscape, but the package 'transparent.sty' is not loaded}%
    \renewcommand\transparent[1]{}%
  }%
  \providecommand\rotatebox[2]{#2}%
  \newcommand*\fsize{\dimexpr\f@size pt\relax}%
  \newcommand*\lineheight[1]{\fontsize{\fsize}{#1\fsize}\selectfont}%
  \ifx\svgwidth\undefined%
    \setlength{\unitlength}{1049.00006104bp}%
    \ifx\svgscale\undefined%
      \relax%
    \else%
      \setlength{\unitlength}{\unitlength * \real{\svgscale}}%
    \fi%
  \else%
    \setlength{\unitlength}{\svgwidth}%
  \fi%
  \global\let\svgwidth\undefined%
  \global\let\svgscale\undefined%
  \makeatother%
  \begin{picture}(1,0.59389893)%
    \lineheight{1}%
    \setlength\tabcolsep{0pt}%
    \put(0,0){\includegraphics[width=\unitlength,page=1]{curvecd_surface_approx.pdf}}%
    \put(0.6104184,0.36485233){\color[rgb]{0,0,0.78431373}\makebox(0,0)[lt]{\lineheight{1.25}\smash{\begin{tabular}[t]{l}\textit{$x$}\end{tabular}}}}%
    \put(0.4696422,0.3726399){\color[rgb]{0,0,0.78431373}\makebox(0,0)[lt]{\lineheight{1.25}\smash{\begin{tabular}[t]{l}\textit{$y$}\end{tabular}}}}%
    \put(0.5476947,0.39323837){\color[rgb]{0,0,0.78431373}\makebox(0,0)[lt]{\lineheight{1.25}\smash{\begin{tabular}[t]{l}\textit{$z$}\end{tabular}}}}%
    \put(-0.00250548,0.00650312){\color[rgb]{0,0,0}\makebox(0,0)[lt]{\lineheight{1.25}\smash{\begin{tabular}[t]{l}\textit{$\{ 0 \}$}\end{tabular}}}}%
    \put(0.09975348,0.00578986){\color[rgb]{0,0,0}\makebox(0,0)[lt]{\lineheight{1.25}\smash{\begin{tabular}[t]{l}\textit{$x$}\end{tabular}}}}%
    \put(0.09592464,0.0914442){\color[rgb]{0,0,0}\makebox(0,0)[lt]{\lineheight{1.25}\smash{\begin{tabular}[t]{l}\textit{$y$}\end{tabular}}}}%
    \put(0.04339306,0.11578667){\color[rgb]{0,0,0}\makebox(0,0)[lt]{\lineheight{1.25}\smash{\begin{tabular}[t]{l}\textit{$z$}\end{tabular}}}}%
    \put(0.49062003,0.2821188){\color[rgb]{0,0,0.78431373}\makebox(0,0)[lt]{\lineheight{1.25}\smash{\begin{tabular}[t]{l}\textit{$\{ A \}$}\end{tabular}}}}%
  \end{picture}%
\endgroup%

%% file: sections/3_surface_mapping.tex
\section{SURFACE MAPPING}
\label{sec:surface_mapping}
The local approximation is facilitated to update the mapping of the surface.
Therefore, a Kalman Filter (KF) estimates the height
\eqref{eq:surface_representation} of each grid point.
To consider the locality of the approximation, the update is weighted by the
distance between the measurement and the grid points.
The proposed approach provides an estimate for the approximation quality by
considering the a-posteriori state covariance of each grid point.
The computational burden is significantly reduced by performing the update only
for grid points in the vicinity of the measurement points using a mask.

\subsection{Kalman Filter Based Height Estimate}
\label{subsec:Kalman-filter_grid}
Because the exact height of the surface in each grid point is unknown, it can
only be estimated as a stochastic variable.
Assume that the height is normal distributed
$
  \zEst_{ij} \sim \mathcal{N}(\zReal_{ij}, \, \sigma_{ij}^2)
$
with mean value $\zReal_{ij}$ and variance $\sigma_{ij}^2$
\cite{grewal_kalman_2015}.

The iterative mapping of the surface, using local geometry approximation, is
based on a discrete linear KF. Each local approximation updates the
current region $\mathcal{A}$ in the global surface representation.
In the following, each grid point is updated by an individual KF with a single
state.
The height at each grid point is estimated by
\begin{align}
  \begin{split}
    &\zEst_{ij}(k | k - 1) = f_{ij} \, \zEst_{ij}(k - 1 | k - 1) \\
    &\covState_{ij}(k | k - 1) = f_{ij}^2 \, \covState_{ij}(k - 1 | k -
    1) + \covProcess_{ij}(k) \\
    &\innovationCov_{ij}(k) = \covApprox_{ij}(k) + h_{ij}^2 \, \covState_{ij}(k
    | k - 1) \\
    &\kalmanGain_{ij}(k) = h_{ij} \, \covState_{ij}(k | k - 1)
    (\innovationCov_{ij}(k))^{-1} \\
    &\zEst_{ij}(k | k) = \zEst_{ij}(k | k - 1) + \kalmanGain_{ij}(k) [
    \zMeasurement_{ij} - h_{ij} \zEst_{ij}(k | k - 1)] \\
    &\covState_{ij}(k | k) = (1 - \kalmanGain_{ij}(k) h_{ij}) \covState_{ij}(k
    | k - 1)
    \label{eq:grid_kalman_filter}
  \end{split}
\end{align}
for the iteration step $k$ with the process covariance $\covProcess_{ij}(k)$ and
the approximation covariance $\covApprox_{ij}(k)$ \cite{kalman_new_1960}.
The measurement $\zMeasurement_{ij}$ is the $z$-coordinate of the local
geometry approximation \eqref{eq:z_measurement} at the grid point with $h_{ij} =
1$.
In case of a time-invariant geometry $f_{ij} = 1$ and $\covProcess_{ij}(k) = 0$
holds for all $i$, $j$ and $k$.

In Section \ref{subsec:local_geometry_approximation}, it is assumed that the
approximation is only fulfilled locally inside the area $\mathcal{A}$.
Therefore, only the grid points inside this area with the indices $(i, \, j) \in
\mathcal{I}$ are updated by \eqref{eq:grid_kalman_filter}.
For all other grid points $\zEst_{ij}(k | k) = \zEst_{ij}(k -1 | k - 1)$ and
$\covState_{ij}(k | k) = \covState_{ij}(k - 1 | k - 1)$ hold.

\subsection{Approximation Covariance Function}
\label{subsec:approximation_covariance_function}
\begin{figure}
  \vspace{-2mm}
  \begin{subfigure}[t]{0.47\linewidth}
    \centering
    \includegraphics{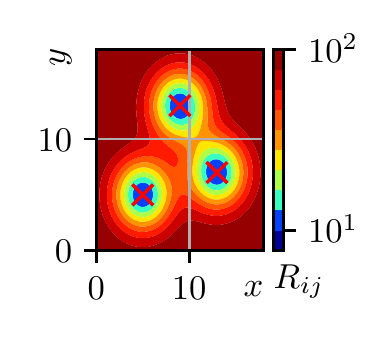}
    \subcaption{Projected on $xy$-plane.}
    \label{subfig:approximation_covariance}
  \end{subfigure}
    \begin{subfigure}[t]{0.47\linewidth}
      \centering
      \includegraphics{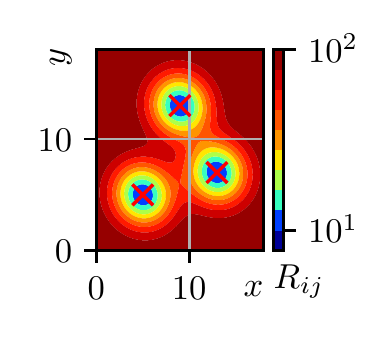}
      \subcaption{Projected on approx. plane.}
      \label{subfig:approximation_covariance_tilted}
    \end{subfigure}
  \caption{Approximation covariance for three measurement points with $\alpha =
  0.1$, $\covApprox_\text{min} = 10$ and $\covApprox_\text{min} = 100$ for the
  example measurement points.}
  \label{fig:approximation_covariance}
\end{figure}

The local approximation provides a measurement $\zMeasurement_{ij}$ for each
grid point.
However, the error may increase further away from the approximated region.
Contrary, we may assume that the approximation is useful in
the vicinity of the measurement points $\point_1, \, \ldots, \, \point_L$, which
describe points on the real surface.
Therefore, we construct a linear combination of RBFs that weights the
approximation update of the grid points, where each measurement point represents
the center of a RBF.
As a heuristic, we assume that the covariance of the local approximation at each
grid point increases with its distance from the $L$ measurement points.
This originates from the consideration that the probability of the approximated
height is larger close to the measurement points.
Here, we choose a sum of RBFs with a Gaussian kernel
\begin{align}
  \begin{split}
    \covApprox_{ij} &= \frac{\covApprox_\text{max} - \covApprox_\text{min}}{L}
    \bigg( 1 - \sum_{l=0}^L e^{-\alpha \| \vec{d}_{ijl} \|^2} \bigg) +
    \covApprox_\text{min} \\ \vec{d}_{ijl} &= \point_l - (\point_l -
    \point_0^S(x_i, \,y_j)) \cdot \vec{n} \; \vec{n} - \point_0^S(x_i, \,y_j)
    \label{eq:approximation_covariance_tilted_plane}
  \end{split}
\end{align}
with the Euclidean norm  $\| \cdot \|$ of the vector, where
$\covApprox_\text{max}$ and $\covApprox_\text{min}$ define the minimum and
maximum covariance value and $\alpha$ how fast the approximation covariance
increases with the distance from the measurement point.
The term $(\point_l - \point_0^S(x_i, \,y_j)) \cdot \vec{n}$ is the distance
between the point and the surface, so that $\| \vec{d}_{ijl} \|$ is the
projected distance on the plane.
For only three measurement points, this distance is equal to the Euclidean
distance because the measurement points lie on the plane.
Another choice may be to project the distance onto the $xy$-plane of $\{0\}$ so
that $\| \vec{d}_{ijl} \| = (x_i - x_l)^2 + (y_j - y_l)^2$.
Also, other kernel functions depending on $\| \vec{d}_{ijl} \|$ can be used.
Figure \ref{fig:approximation_covariance} shows the approximation covariance
projected onto the approximation plane and $xy$-plane for the measurement points
$\point_1 = [5, \, 5, \, 5]\transp$, $\point_2 = [13, \, 7, \, 7]\transp$ and
$\point_3 = [9, \, 13, \, 10]\transp$.

Determining the approximation covariance for the whole grid is very
time-consuming.
Therefore, the calculation is only performed for the grid point at the indices
$(i, \, j) \in \mathcal{I}$.

\subsection{Masked Update}
\label{subsec:masked_update}
To lower the computational effort and avoid undesired changes outside the
approximated region, it is desired to update the mapping only inside the region
$\mathcal{A}$.
This area is defined by the position of the measurement points on the surface
and may have various shapes.
In the following, we define index sets $\mathcal{I}$ to construct different
binary masks with the elements
\begin{equation}
  \mask_{ij} =
  \bigg\{\begin{array}{ll}
    1 & (i, \, j) \in \mathcal{I}\\
    0 & \text{else}
  \end{array}
  \label{eq:mask_definitions}
\end{equation}
of an $M \times N$ matrix.
\subsubsection{Region of interest (ROI) mask}
The ROI mask is a rectangular mask, whose size is defined by the minimum and
maximum $x$- and $y$-coordinates.
Therefore, the elements of the index set $\mathcal{I}$ in \eqref{eq:mask_definitions} fulfill
\begin{equation*}
  \mathcal{I} := \lbrace (i, \, j) \vert \,
    x_i \geq x_\text{min} \; \land \; x_i \geq x_\text{max}
    \; \land \;
    y_j \leq y_\text{min} \; \land \; y_j \leq y_\text{max}
    \rbrace
\end{equation*}
with the minimal $x$-coordinate $x_\text{min} = \min \{ p_{1, \, x}, \,
\ldots, \, p_{L, \, x} \}$ of the measurement points and maximal $x$-
or $y$-coordinate $x_\text{max} = \max \{ p_{1, \, x}, \, \ldots, \, p_{L, \, x}
\}$ respectively.

\subsubsection{Triangular mask}
The triangular mask directly gives the convex area between three measurement
points.
It is defined by considering the triangle edges as
\begin{align*}
  \begin{split}
  &a_1 = (y_j - p_{1, \, y}) (p_{2, \, x} - p_{1, \, x}) - (p_{2, \, y} - p_{1,
  \, y}) (x_i - p_{1, \, x}) \\
  &a_2 = (y_j - p_{2, \, y}) (p_{3, \, x} - p_{2, \, x}) - (p_{3, \, y} - p_{2,
  \, y}) (x_i - p_{2, \, x}) \\
  &a_3 = (y_j - p_{3, \, y}) (p_{1, \, x} - p_{3, \, x}) - (p_{1, \, y} - p_{3,
  \, y}) (x_i - p_{3, \, x}) \\
  &\mathcal{I} := \lbrace (i, \, j) \vert \, a_1  \geq 0 \; \land \; a_2 \geq 0 \; \land \; a_3 \geq 0 \rbrace,
  \end{split}
\end{align*}
where $p_{l, \, x}$ and $p_{l, \, y}$ with $l \in \{1, \, 2, \, 3\}$ are the
$x$- and $y$-coordinates of the measurement points.

\subsubsection{Largest circle mask}
This circular mask includes all measurement points by choosing the mean point
$\point_c$ from \eqref{eq:measurement_points_covariance} as the center point of
the circle and the radius as the largest distance between this point and all
measurement points. Therefore the index set can be written as
\begin{equation}
      \mathcal{I} := \lbrace (i, \, j) \vert \, (x_i - p_{c, \, x})^2 + (y_j -
      p_{c, \, y})^2 - \radius \geq 0 \rbrace,
      \label{eq:circular_mask}
\end{equation}
where $ p_{c, \, x}$ and $ p_{c, \, y}$ are the $x$- and $y$-coordinates of the
point $\point_c$.
The radius is chosen as $\radius = \max \{ r_l \vert \, r_l = \sqrt{(p_{l, \, x}
- p_{c, \, x})^2 + (p_{l, \, y}- p_{c, \, y})^2}, \, \forall l \}$ by projecting the
measurement points onto the $xy$-plane of frame $\{ 0 \}$ or $\{ A \}$.

\subsubsection{Circle around points (CAP) mask}
Another option is to create a circular mask around each measurement point
$\point_l = [p_{l, \, x}, \; p_{l, \, y}, \; p_{l, \, z}]\transp$ for
$l \in \{ 1, \, \ldots, \, L \}$ with a fixed radius $r$.
This is achieved by exploiting \eqref{eq:circular_mask} to
\begin{equation*}
  \mathcal{I} := \bigg\lbrace (i, \, j) \bigg\vert \bigvee_{l = 1}^L (x_i -
  p_{l, \, x})^2 + (y_j - p_{l, \, y})^2 - \radius \geq 0 \bigg\rbrace
\end{equation*}
concatenating single circular masks using the logical or operator.
The mask allows additional tuning of the radius.

\begin{figure}
  \vspace{-2mm}
  \begin{subfigure}[t]{0.47\linewidth}
      \centering
      \includegraphics{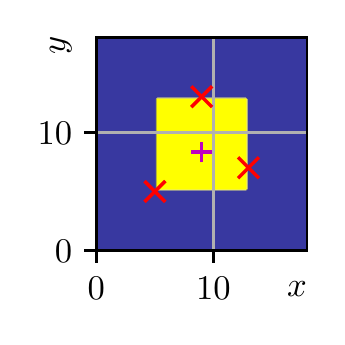}
      \subcaption{Rectangular ROI mask.}
      \label{subfig:mask_roi}
  \end{subfigure}
  \hfill %
  \begin{subfigure}[t]{0.47\linewidth}
      \centering
      \includegraphics{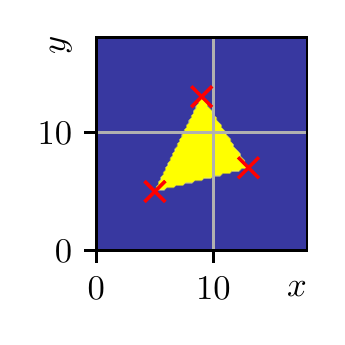}
      \subcaption{Triangle mask.}
      \label{subfig:mask_triangle}
  \end{subfigure}
  \\
  \begin{subfigure}[t]{0.47\linewidth}
      \centering
      \includegraphics{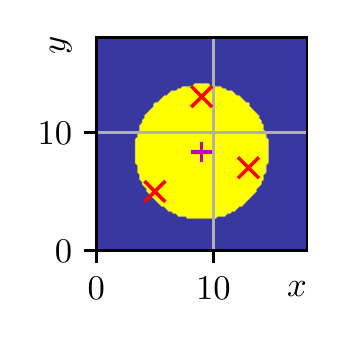}
      \subcaption{Largest circle mask.}
      \label{subfig:mask_largest_circle}
  \end{subfigure}
  \hfill %
  \begin{subfigure}[t]{0.47\linewidth}
    \centering
    \includegraphics{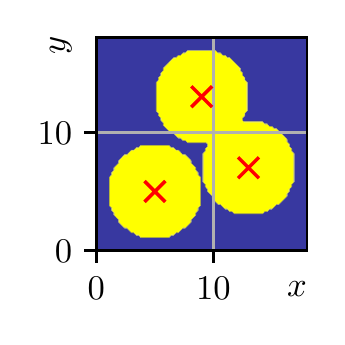}
    \subcaption{Circle around points mask.}
    \label{subfig:mask_circle_around_points_py}
  \end{subfigure}
  \caption{Different mask types projected onto the approximation plane for the
  example measurement points.}
  \label{fig:masks}
\end{figure}

The described masks are shown in Fig. \ref{fig:masks}.
They are determined in the local frame $\{ \text{A} \}$ by transforming the
measurement and approximation points using \eqref{eq:rot_mat_local_frame} to
consider the local geometry, similar to
\eqref{eq:approximation_covariance_tilted_plane}.
It is also possible to enlarge the boundaries of the mask, e.g., by using the
dilation operator, which is known from image processing
\cite{gonzalez_digital_2018}.

%% file: sections/4_experimental_setup.tex
\section{EXPERIMENTAL SETUP}
\label{sec:experimental_setup}
\begin{figure}
  \vspace{1.5mm}
  \centering
  \includegraphics[width=75mm]{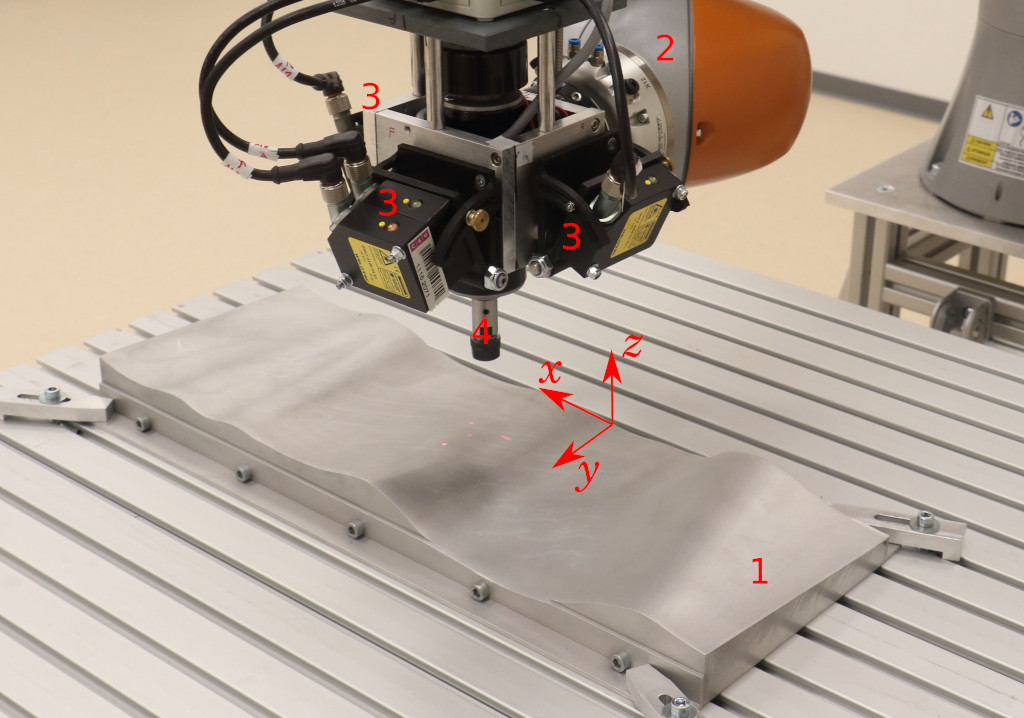}
  \caption{Experimental setup with freeform surface (1) and robotic
  manipulator (2). The end-effector is equipped with three laser distance
  sensors (3) and an electric spindle (4).}
  \label{fig:experimental_setup}
\end{figure}

The proposed method is tested by mapping an example surface of a computerized
numerical control (CNC) manufactured aluminum part with the dimensions $500
\times 200 \times 50 \, \si{mm}$ and a height difference of max. $\SI{30}{\mm}$.
The surface has a minimal curvature radius of $\radius_\text{min} =
1/\kappa_\text{max} = \SI{20}{\mm}$.
The surface approximation is performed using three laser distance sensors of
type \emph{Welotec OWLF 4030 FA S1} with a measurement range between $50$ and
$\SI{300}{\mm}$.
The sensors have a resolution of $\SI{0.33}{\mm}$ and a maximum linearization
error of $\SI{1}{\mm}$ over the full measurement range.
They are mounted on a \emph{KUKA LBR iiwa 14} robotic manipulator.
Together with the forward kinematics of the manipulator, the coordinates of the
measured points on the surface are determined.
The sensors are tilted by $\SI{30}{\degree}$ to achieve a smaller approximation
area in the vicinity of the tool.
The exact tilting angle is identified by a calibration routine using a linear
regression of the distance measurements.
The analog signals of the distance sensors are evaluated on a
\emph{Beckhoff CX2000} programmable logic controller (PLC).
Figure \ref{fig:experimental_setup} shows the example surface with the laser
distance sensors mounted on the robotic tool.

The surface distance and orientation tracking controller proposed in
\cite{amersdorfer_real-time_2020} is used to track the a-priori unknown geometry
of the surface.
The controller runs on the PLC using the current end-effector pose and the
distance measurements to determine the desired pose.
This pose is transferred to the \emph{KUKA Sunrise} robot controller via
\emph{EtherCAT}, where it is continuously commanded to the robotic manipulator
using the \emph{KUKA Sunrise.Servoing} soft real-time interface.

The proposed mapping method is implemented in C and executed as \emph{MATLAB
MEX} file and runs on a separate PC\footnote{Intel Core i5-7200U CPU
2.50GHz$\times$4, 24GiB RAM, Linux 5.10.19-200.fc33.x86\_64, MATLAB 2020b.}.
The algorithm is parallelized using \emph{OpenMP}, which allows update times of
$\SI{1.5}{\ms}$ for the used $\SI{2}{\mm}$ grid step.
It can be either used offline with stored measurements or online where
the measured points are received directly from the PLC.

%% file: sections/5_results_and_discussion.tex
\begin{figure}
  \vspace{-2mm}
  \begin{subfigure}[t]{\linewidth}
    \centering
    \includegraphics{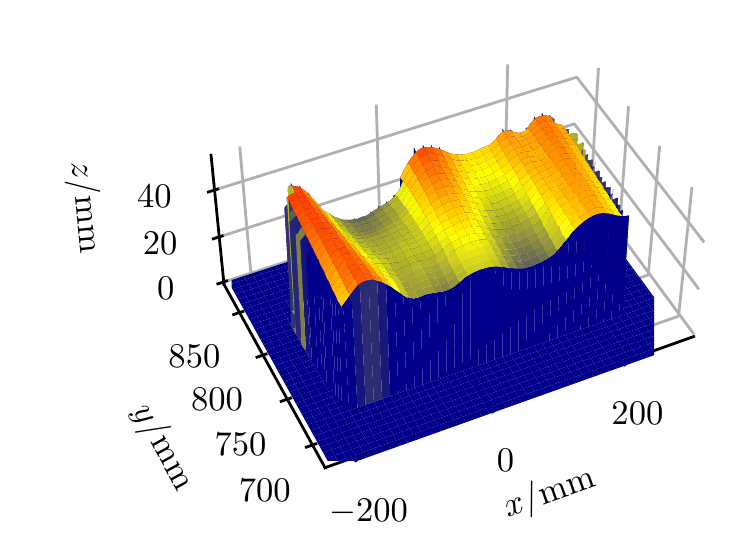}
    \vspace{4mm}
    \subcaption{Mapped surface.}
    \label{subfig:surface_mapping_const_height}
  \end{subfigure}
  \begin{subfigure}[t]{\linewidth}
    \centering
    \includegraphics{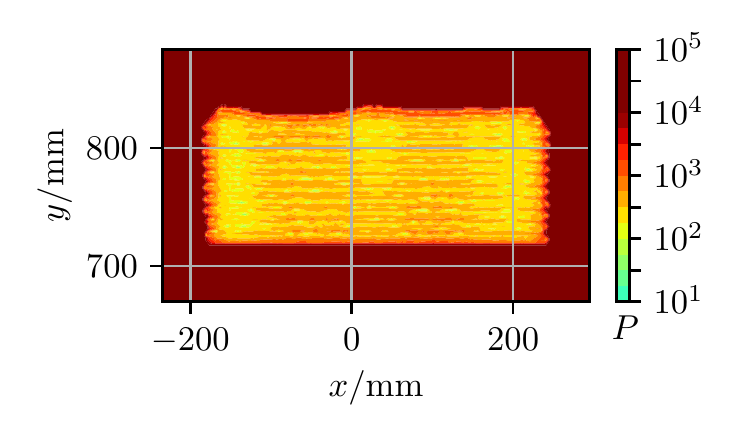}
    \subcaption{Mapping covariance.}
    \label{subfig:mapping_covariance_const_height}
  \end{subfigure}
  \begin{subfigure}[t]{\linewidth}
    \centering
    \includegraphics{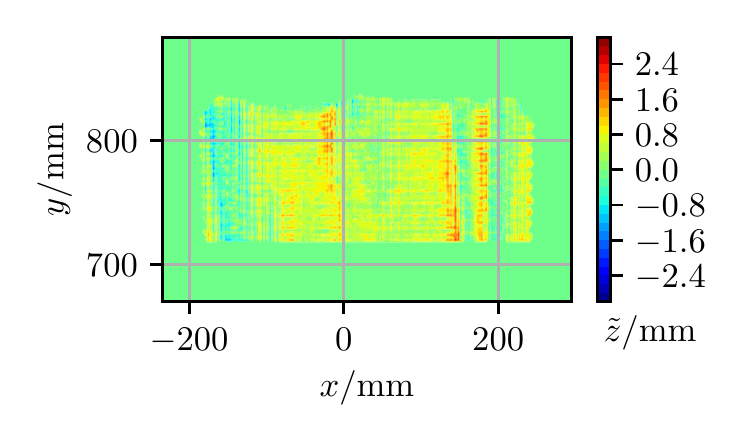}
    \subcaption{Mapping error.}
    \label{subfig:mapping_error_const_height}
  \end{subfigure}
  \caption{Mapping results of the constant height trajectory.
  }
  \label{fig:surface_tracking_mapping_const_height}
\end{figure}
\begin{figure}
  \vspace{-2mm}
  \begin{subfigure}[t]{\linewidth}
    \centering
    \includegraphics{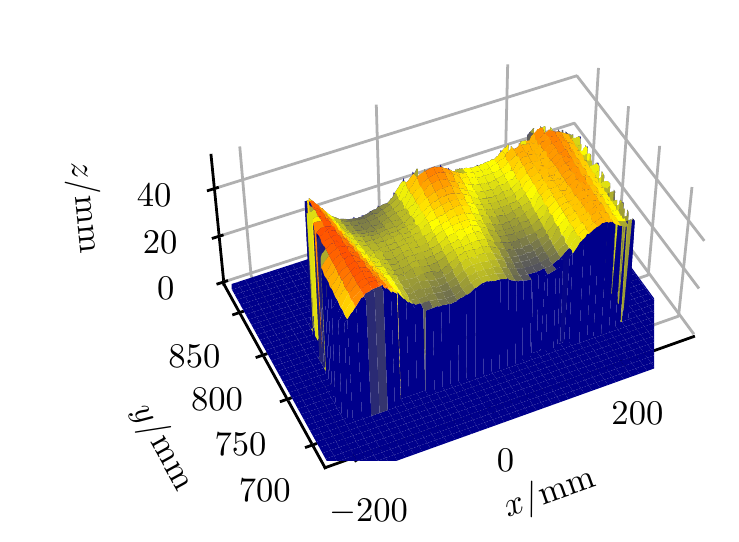}
    \vspace{4mm}
    \subcaption{Mapped surface.}
    \label{subfig:surface_mapping_surface_tracking}
  \end{subfigure}
  \begin{subfigure}[t]{\linewidth}
    \centering
    \includegraphics{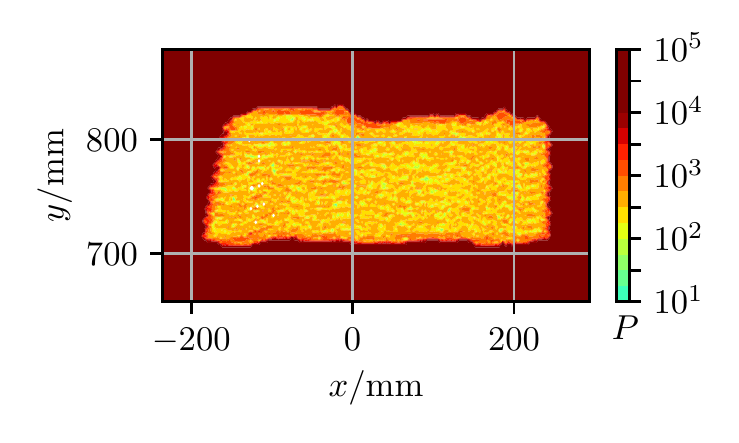}
    \subcaption{Mapping covariance.}
    \label{subfig:mapping_covariance_surface_tracking}
  \end{subfigure}
  \begin{subfigure}[t]{\linewidth}
    \centering
    \includegraphics{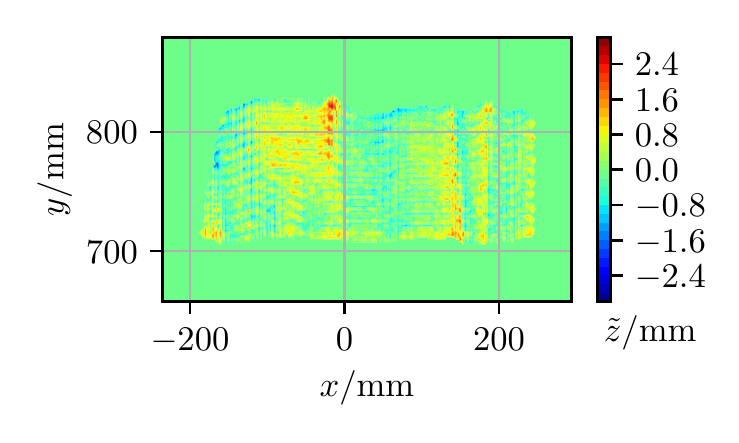}
    \subcaption{Mapping error.}
    \label{subfig:mapping_error_surface_tracking}
  \end{subfigure}
  \caption{Mapping results of the surface tracking trajectory.
  }
  \label{fig:surface_tracking_mapping_surface_tracking}
\end{figure}
\section{RESULTS AND DISCUSSION}
\label{sec:results_and_discussion}
The proposed methods are evaluated on the experimental setup described above by mapping the surface of the aluminum part.
Here, the robotic manipulator is used to move the distance sensors along the surface.
A path is planned with $21$ parallel lines with a distance of $\SI{5}{\mm}$ and
alternating directions to cover a rectangular area of $400 \times \SI{100}{\mm}$
on the surface.
It is executed with a velocity of $\SI{25}{\mm\per\s}$.

The experiments are performed with the masks defined in Section
\ref{subsec:masked_update}.
For the CAP mask, the radius is set to $\SI{5}{mm}$.
To satisfy a coverage of the surface, even when the approximation area is small,
the mask is enlarged using the dilate operation with two grid steps.
The parameter $\alpha = 0.1$ of the approximation covariance is chosen and
the minimum and maximum covariance are set to $\covApprox_\text{min} = 10^1$ and
$\covApprox_\text{max} = 10^4$.
This ensures that the approximation covariance functions close to each other
overlap.

The mapping update is executed when the distance to the mean point of the last
update exceeds $\SI{2}{\mm}$.
This avoids over-fitting of the approximation plane when the end-effector does
not move and ensures that the area covered between two updates is sufficiently
large.
The measurements between the updates can be used to improve the local
approximation.

To evaluate the quality of the mapping, the height $\zReal_{ij}$ of the example
surface is sampled with one distance sensor at the end-effector, pointing
anti-parallel to the $z$-axis of frame $\{ 0 \}$.
For the sample points, an equidistant grid of $\SI{5}{\mm}$ in the $xy$-plane of
the inertial coordinate system is used.
The errors are computed by using the mean and maximum absolute value of error
$\zError_{ij} = \zEst_{ij} - \zReal_{ij}$ between the sample points and the
mapping at the $x$- and $y$-coordinates closest to these points.
Therefore, only grid points with $\covState_{ij} \leq 10^4$ are considered.
\begin{table}
  \caption{Approximation error for different mask types for the constant height and surface tracking trajectory in $\si{\mm}$.}
  \label{tab:approximation_error_tilted}
  \centering
  \begin{tabular}{l|l|c|c|c}
    Experiment & Mask & Mean Err. & Max. Err. & Std. Dev. \\ %
    \hline
    const. height
    & triangle & 0.467 & 2.073 & 0.324 \\ %
    & largest circle & 0.556 & 2.978 & 0.414 \\ %
    & CAP & 0.534 & 1.991 & 0.295 \\ %
    & ROI & 0.694 & 4.497 & 0.522 \\ %
    \hline
    surf. tracking
    & triangle & 0.443 & 2.554 & 0.345 \\ %
    & largest circle & 0.468 & 3.650 & 0.436 \\ %
    & CAP & 0.440 & 3.425 & 0.402 \\ %
    & ROI & 0.572 & 4.364 & 0.521  %
  \end{tabular}
\end{table}

\subsection{Constant Height Trajectory}
\label{subsec:constant_height_trajectory}
In the first experiment, the end-effector tracks the planned path at a constant
height over the surface to investigate the influence of approximation area size.
Due to the tilt angle of the sensors, the size of the approximation
area changes depending on the distance between the surface and the tool.

The mapped surface is shown in Fig. \ref{subfig:surface_mapping_const_height}
based on the triangular mask with dilation.
All grid points that are not mapped remain zero.
The mapping reconstructs the geometry of the example surface shown in Fig.
\ref{fig:experimental_setup} well.
In Tab. \ref{tab:approximation_error_tilted}, the mean and maximum
absolute mapping errors as well as the error standard deviation, are given for
the different mask types described in Section \ref{subsec:masked_update}.
The CAP mask has the smallest maximum error, while the
triangular mask shows the smallest mean error.
This may come from the fact that the planar approximation with three measurement
points provides a reasonable estimate for the heights in the vicinity of the
measurement points as long as the local curvature is small.
The triangular mask only contains points, which are between the measurement
points and also covers the smallest area of all masks.
Therefore, the local approximation does not influence points outside this area.

Figure \ref{subfig:mapping_covariance_const_height} depicts the state covariance
of each grid point.
There, the contour of the mapped area is clearly visible.
Since two laser distance sensor measurements are always performed at the same
$y$-coordinate, their paths can be recognized.
Because of the constant height of the end-effector and the directions of the
distance sensors pointing to a common point, the distance between the
measurements decreases for lower areas of the surface.
This leads to smaller approximation areas and, therefore, locally smaller
approximation covariances.
Therefore, also the state covariance is reduced.
The area covered by the triangular mask varies between $88$ and
$\SI{420}{\square\mm}$, with a mean value of $\SI{221}{\square\mm}$.

Due to the local planar approximation of the surface, the mapping error, shown
in Fig. \ref{subfig:mapping_error_const_height}, is mainly affected by the
curvature of the surface.
For areas where the curvature is convex, the surface height is underestimated,
while in concave areas, it is overestimated.
Because of the constant height trajectory, the distance between the measurement
points increases for higher surface areas, leading to a larger approximation
error in regions with similar curvature.
Using a coaxial alignment of the sensors or keeping a constant distance between
tool and surface can reduce this effect.

\subsection{Surface Tracking Trajectory}
\label{subsec:surface_tracking_trajectory}
In the second experiment, the end-effector follows the surface geometry using
the distance and orientation controller with a constant distance between
end-effector and surface.
This leads to the fact that the approximation area remains nearly constant.
For example, the triangular mask covers an area between $52$ and
$\SI{106}{\square\mm}$, with a mean value of $\SI{82}{\square\mm}$.
This is $\SI{36}{\percent}$ smaller as for the constant height trajectory.

The main deviations between Fig. \ref{fig:surface_tracking_mapping_const_height}
and Fig. \ref{fig:surface_tracking_mapping_surface_tracking} originate from the
different scanning paths because of the distance and orientation tracking
controller.
The mapped heights are shown in Fig.
\ref{subfig:surface_mapping_surface_tracking}.
While the geometry, in general, is also approximated accurately, it can be
noticed that the surface is not as smooth as in the constant height trajectory
experiment above.
Also, the results in Tab. \ref{tab:approximation_error_tilted} show larger
maximum errors, while the mean errors are smaller.
The reason for this may be that the motion along the unknown surface is not
smooth anymore, because the distance and orientation controller must
continuously adapt the end-effector pose.
Together with some minor but different delays in the measurements of the
distance sensors and the end-effector pose, this leads to errors in the
estimation of the approximation plane.
As shown in Fig. \ref{subfig:mapping_error_surface_tracking}, the error
is independent of the current height of the surface, and the local
curvature has a smaller influence due to the small approximation area.
The state covariance depicted in Fig.
\ref{subfig:mapping_covariance_surface_tracking} also shows that the estimate
does not depend on the surface height.
Without the dilation of the mask, the surface is not completely mapped, such
that some grid points are never updated.
This would lead to holes inside the mapping.

During the surface tracking trajectory along the unknown geometry also machining
tasks like polishing, grinding or cleaning can be performed.
After the surface is covered once, the mapping provides an estimate of the
current surface geometry.
This can be used to refine the path and trajectory planning for the next
execution of the process.
In this next execution, the mapping is updated again to improve the estimate of
the geometry iteratively.
This is especially useful for machining processes, which are repeated multiple
times on the same part.
The approach is not limited to laser distance sensors such that other types of
distance sensors can be used.
The mean error, given in Tab. \ref{tab:approximation_error_tilted}, is one order
of magnitude larger than commercial structured light 3D scanners with an accuracy down to $\SI{30}{\micro\m}$ \cite{mendricky_accuracy_2020}  for a similar measurement volume but is in the range of the accuracy of the
robotic manipulator \cite{besset_advanced_2016}.
Therefore, the accuracy of the proposed mapping method is sufficient for the
planning of force or impedance-controlled robotic machining tasks or in applications where the manipulator equipped with a compliant end-effector. However, for position controlled tasks with a stiff tool the risk of undesired collisions may occur.
The approach can be applied to all industrial robots providing real-time
kinematic measurements.

%% file: sections/6_conclusion.tex
\section{CONCLUSION AND FUTURE WORK}
\label{sec:conclusion}
The paper proposes a method to map a curved freeform surface using a local
planar approximation for scenarios, where only sparse measurements are
available, but sensors can be moved along the surface.
Linear Kalman filters are used to estimate the height at each grid point.
The local measurements are weighted using radial basis functions with a Gaussian
kernel to estimate the approximation covariance of the grid points.
To update only areas where the approximation is valid different mask types are
proposed.
The mapping method is evaluated by performing two experiments on an example
surface.
Here, the robotic manipulator moves the tool equipped with three distance
sensors over the surface.
Once at a constant height and once by tracking the surface geometry with a
distance and orientation controller.
Both experiments show that the surface is mapped accurately, while the mapping
error depends primarily on the size of the approximation area and the curvature
of the surface.

Because the updates of the grid points are independent, the mapping method is
well suited for massive parallelization on GPUs.
This allows integration into online path planning applications in future
works.
It is also intended to extend the approach to a SLAM method to improve both
the mapping accuracy and the estimate for the tool pose.
Another approach is to use higher-order approximations of the local surface
geometry, e.g., paraboloid or NURBS, instead of the plane as a first-order
approximation.
This may require more sensors or can be achieved by exploiting the iterative
nature of the method using the previous estimate of the surface geometry.

%% file: root.bbl
\begin{thebibliography}{10}
\providecommand{\url}[1]{#1}
\csname url@samestyle\endcsname
\providecommand{\newblock}{\relax}
\providecommand{\bibinfo}[2]{#2}
\providecommand{\BIBentrySTDinterwordspacing}{\spaceskip=0pt\relax}
\providecommand{\BIBentryALTinterwordstretchfactor}{4}
\providecommand{\BIBentryALTinterwordspacing}{\spaceskip=\fontdimen2\font plus
\BIBentryALTinterwordstretchfactor\fontdimen3\font minus
  \fontdimen4\font\relax}
\providecommand{\BIBforeignlanguage}[2]{{%
\expandafter\ifx\csname l@#1\endcsname\relax
\typeout{** WARNING: IEEEtran.bst: No hyphenation pattern has been}%
\typeout{** loaded for the language `#1'. Using the pattern for}%
\typeout{** the default language instead.}%
\else
\language=\csname l@#1\endcsname
\fi
#2}}
\providecommand{\BIBdecl}{\relax}
\BIBdecl

\bibitem{ren_curve_2017}
M.~Ren, L.~Kong, L.~Sun, and C.~Cheung, ``A {Curve} {Network} {Sampling}
  {Strategy} for {Measurement} of {Freeform} {Surfaces} on {Coordinate}
  {Measuring} {Machines},'' \emph{IEEE Transactions on Instrumentation and
  Measurement}, vol.~66, no.~11, pp. 3032--3043, Nov. 2017.

\bibitem{hansen_structured_2014}
K.~Hansen, J.~Pedersen, T.~Sølund, H.~Aanæs, and D.~Kraft, ``A {Structured}
  {Light} {Scanner} for {Hyper} {Flexible} {Industrial} {Automation},'' in
  \emph{2014 2nd {International} {Conference} on {3D} {Vision}}, vol.~1, Dec.
  2014, pp. 401--408, iSSN: 1550-6185.

\bibitem{hennad_characterization_2019}
A.~Hennad, P.~Cockett, L.~McLauchlan, and M.~Mehrubeoglu, ``Characterization of
  {Irregularly}-{Shaped} {Objects} {Using} {3D} {Structured} {Light}
  {Scanning},'' in \emph{2019 {International} {Conference} on {Computational}
  {Science} and {Computational} {Intelligence} ({CSCI})}, Dec. 2019, pp.
  600--605.

\bibitem{song_distortion-free_2019}
D.~Song and Y.~J. Kim, ``Distortion-free {Robotic} {Surface}-drawing using
  {Conformal} {Mapping},'' in \emph{2019 {International} {Conference} on
  {Robotics} and {Automation} ({ICRA})}, May 2019, pp. 627--633, iSSN:
  2577-087X.

\bibitem{chu_visible_1988}
C.-C. Chu and A.~C. Bovik, ``\BIBforeignlanguage{en}{Visible surface
  reconstruction via local minimax approximation},''
  \emph{\BIBforeignlanguage{en}{Pattern Recognition}}, vol.~21, no.~4, pp.
  303--312, Jan. 1988.

\bibitem{zhao_development_2008}
Y.~Zhao, J.~Zhao, L.~Zhang, and L.~Qi, ``Development of a {Robotic} {3D}
  {Scanning} {System} for {Reverse} {Engineering} of {Freeform} {Part},'' in
  \emph{2008 {International} {Conference} on {Advanced} {Computer} {Theory} and
  {Engineering}}, Dec. 2008, pp. 246--250, iSSN: 2154-7505.

\bibitem{ganesh_versatile_2012}
G.~Ganesh, N.~Jarrassé, S.~Haddadin, A.~Albu-Schaeffer, and E.~Burdet, ``A
  versatile biomimetic controller for contact tooling and haptic exploration,''
  in \emph{2012 {IEEE} {International} {Conference} on {Robotics} and
  {Automation}}, May 2012, pp. 3329--3334, iSSN: 1050-4729.

\bibitem{song_artistic_2018}
D.~Song, T.~Lee, and Y.~J. Kim, ``Artistic {Pen} {Drawing} on an {Arbitrary}
  {Surface} {Using} an {Impedance}-{Controlled} {Robot},'' in \emph{2018 {IEEE}
  {International} {Conference} on {Robotics} and {Automation} ({ICRA})}, May
  2018, pp. 4085--4090, iSSN: 2577-087X.

\bibitem{mazzini_tactile_2009}
F.~Mazzini, D.~Kettler, S.~Dubowsky, and J.~Guerrero, ``Tactile robotic mapping
  of unknown surfaces: an application to oil well exploration,'' in \emph{2009
  {IEEE} {International} {Workshop} on {Robotic} and {Sensors} {Environments}},
  Nov. 2009, pp. 80--85.

\bibitem{mazzini_tactile_2011}
F.~Mazzini, D.~Kettler, J.~Guerrero, and S.~Dubowsky, ``Tactile {Robotic}
  {Mapping} of {Unknown} {Surfaces}, {With} {Application} to {Oil} {Wells},''
  \emph{IEEE Transactions on Instrumentation and Measurement}, vol.~60, no.~2,
  pp. 420--429, Feb. 2011.

\bibitem{ozog_real-time_2013}
P.~Ozog and R.~M. Eustice, ``Real-time {SLAM} with piecewise-planar surface
  models and sparse {3D} point clouds,'' in \emph{2013 {IEEE}/{RSJ}
  {International} {Conference} on {Intelligent} {Robots} and {Systems}}, Nov.
  2013, pp. 1042--1049, iSSN: 2153-0866.

\bibitem{hong_three-dimensional_2019}
S.~Hong and J.~Kim, ``\BIBforeignlanguage{en}{Three-{Dimensional} {Visual}
  {Mapping} of {Underwater} {Ship} {Hull} {Surface} using {View}-based
  {Piecewise}-{Planar} {Measurements}},''
  \emph{\BIBforeignlanguage{en}{IFAC-PapersOnLine}}, vol.~52, no.~21, pp.
  384--389, Jan. 2019.

\bibitem{slatton_fusing_2001}
K.~Slatton, M.~Crawford, and B.~Evans, ``Fusing interferometric radar and laser
  altimeter data to estimate surface topography and vegetation heights,''
  \emph{IEEE Transactions on Geoscience and Remote Sensing}, vol.~39, no.~11,
  pp. 2470--2482, Nov. 2001.

\bibitem{faria_probabilistic_2010}
D.~R. Faria, R.~Martins, J.~Lobo, and J.~Dias, ``Probabilistic representation
  of {3D} object shape by in-hand exploration,'' in \emph{2010 {IEEE}/{RSJ}
  {International} {Conference} on {Intelligent} {Robots} and {Systems}}, Oct.
  2010, pp. 1560--1565, iSSN: 2153-0866.

\bibitem{amersdorfer_real-time_2020}
M.~Amersdorfer, J.~Kappey, and T.~Meurer, ``\BIBforeignlanguage{en}{Real-time
  freeform surface and path tracking for force controlled robotic tooling
  applications},'' \emph{\BIBforeignlanguage{en}{Robotics and
  Computer-Integrated Manufacturing}}, vol.~65, p. 101955, Oct. 2020.

\bibitem{mitra_estimating_2003}
N.~J. Mitra and A.~Nguyen, ``Estimating surface normals in noisy point cloud
  data,'' in \emph{Proceedings of the nineteenth annual symposium on
  {Computational} geometry}, ser. {SCG} '03.\hskip 1em plus 0.5em minus
  0.4em\relax New York, NY, USA: Association for Computing Machinery, Jun.
  2003, pp. 322--328.

\bibitem{fransens_hierarchical_2006}
J.~Fransens and F.~Van~Reeth, ``Hierarchical {PCA} {Decomposition} of {Point}
  {Clouds},'' in \emph{Third {International} {Symposium} on {3D} {Data}
  {Processing}, {Visualization}, and {Transmission} ({3DPVT}'06)}, Jun. 2006,
  pp. 591--598.

\bibitem{grewal_kalman_2015}
M.~S. Grewal and A.~P. Andrews, \emph{Kalman filtering: theory and practice
  using {MATLAB}}, fourth edition~ed.\hskip 1em plus 0.5em minus 0.4em\relax
  Hoboken, New Jersey: John Wiley \& Sons Inc, 2015.

\bibitem{kalman_new_1960}
R.~E. Kalman, ``\BIBforeignlanguage{en}{A {New} {Approach} to {Linear}
  {Filtering} and {Prediction} {Problems}},''
  \emph{\BIBforeignlanguage{en}{Journal of Basic Engineering}}, vol.~82, no.~1,
  pp. 35--45, Mar. 1960, publisher: American Society of Mechanical Engineers
  Digital Collection.

\bibitem{gonzalez_digital_2018}
R.~C. Gonzalez and R.~E. Woods, \emph{Digital image processing}.\hskip 1em plus
  0.5em minus 0.4em\relax New York, NY: Pearson, 2018.

\bibitem{mendricky_accuracy_2020}
R.~Mendricky and J.~Sobotka, ``Accuracy {Comparison} of the {Optical} {3D}
  {Scanner} and {CT} {Scanner},'' \emph{Manufacturing Technology}, vol.~20,
  no.~6, pp. 791--801, Dec. 2020.

\bibitem{besset_advanced_2016}
P.~Besset, A.~Olabi, and O.~Gibaru, ``Advanced calibration applied to a
  collaborative robot,'' in \emph{2016 {IEEE} {International} {Power}
  {Electronics} and {Motion} {Control} {Conference} ({PEMC})}, Sep. 2016, pp.
  662--667.

\end{thebibliography}
